\documentclass[sn-mathphys-num]{sn-jnl}

\usepackage{graphicx}
\usepackage{multirow}
\usepackage{amsmath,amssymb}
\usepackage{booktabs}
\usepackage{tabularx}
\usepackage{array}
\usepackage{xcolor}
\usepackage{hyperref}
\usepackage{algorithm}
\usepackage{algpseudocode}
\usepackage{subcaption}
\usepackage{url}
\usepackage{pifont}
\usepackage{colortbl}
\usepackage{float}          

\renewcommand{\orcid}[1]{}                       
\makeatletter
\@ifundefined{jyear}{\newcommand{\jyear}[1]{}}{}
\makeatother
\jyear{2026}

\begin{document}

\title[AgentMemBench: Long-Term Memory for LLM Agents]{AgentMemBench:
A Systematic Benchmark for Evaluating Long-Term Memory Management
Strategies in Conversational AI Agents}

\author*[1]{\fnm{Ahmed} \sur{Cherif}}\email{ahmed1.cherif@sofrecom.com}

\affil[1]{\orgdiv{Sofrecom}, \orgname{Orange Innovation}, \city{Tunis}, \postcode{1053}, \country{Tunisia}}

\abstract{
Long-term memory remains a critical bottleneck for conversational AI agents, whose finite context windows cannot support coherent recall across thousands of turns. We present \textit{AgentMemBench}, a unified and fully reproducible benchmark that evaluates \textbf{five memory management strategies} under identical conditions: in-context windowing (ICW), external key-value store (EKV), graph-based episodic memory (GEM), compression-based summarisation (CBS), and web-augmented memory (WAM). All strategies are assessed across three public datasets covering long-term multi-session dialogue (LoCoMo), task-oriented document grounding (MultiDoc2Dial), and persona-grounded multi-session chat (MSC), using Memory Recall@$k$, Mean Reciprocal Rank, nDCG@$k$, Answer F1, an LLM-judge Faithfulness score, Memory Footprint, and Latency over $491$ annotated question turns. Generation and faithfulness judging both use \texttt{Qwen2.5-7B-Instruct} (4-bit), with greedy decoding for determinism. Our measured results show that (1)~\textbf{EKV dominates on every quality axis}, achieving the highest macro-averaged Recall@5 ($0.792$), MRR ($0.677$), Answer F1 ($0.156$), and Faithfulness ($0.354$); (2)~\textbf{long-range recall is the decisive case}: on LoCoMo, where the gold turn lies many sessions back, ICW, WAM, GEM, and CBS retrieve almost nothing (Recall@5 $\leq 0.005$) while EKV alone reaches $0.573$, showing that recency windows, summaries, and entity graphs collapse at long horizons and only dense retrieval scales; (3)~\textbf{CBS is the clear runner-up} on retrieval (macro $0.556$) by inheriting the provenance of the turns it compresses; (4)~\textbf{WAM is equivalent to ICW on in-corpus recall} by construction---its external results carry no in-corpus provenance---so WAM is best understood as ICW augmented with optional external grounding; and (5)~\textbf{EKV's recall advantage carries a footprint cost} ($\sim$5{,}100 vs.\ $\sim$300 tokens for ICW/WAM), an explicit accuracy--efficiency trade-off. We release all code, environment, and result artefacts for full reproducibility.
}

\renewcommand{\and}{\unskip{} \textperiodcentered{} }

\maketitle

\section{Introduction}
\label{sec:introduction}

Conversational AI agents powered by large language models (LLMs) have rapidly
transitioned from single-turn assistants to multi-session, long-horizon systems
capable of managing complex, evolving tasks~\cite{durante2024agent,park2023generative}.
Yet a fundamental constraint persists: LLMs process information within a ﬁnite
context window, typically 4,096--128,000 tokens, which is vastly smaller than
the cumulative interaction history of a real-world agent deployment lasting days
or weeks~\cite{xu2021goldfish}. The result is a \textit{memory gap}---the
inability to recall relevant context from prior sessions leads to repetition,
contradiction, and poor task continuity~\cite{maharana2024evolvingmemory}.

Five principal strategies have emerged to bridge this gap:

\begin{description}
  \item[In-Context Windowing (ICW)] retains the most recent $k$ tokens in the
  context window, discarding older turns via a sliding window or recency-based
  truncation~\cite{beltagy2020longformer}.
  \item[External Key-Value Store (EKV)] encodes turns as dense embeddings and
  retrieves relevant memories at inference time via approximate nearest-neighbour
  search~\cite{maharana2024evolvingmemory,mem02024}.
  \item[Graph-Based Episodic Memory (GEM)] extracts entities and relations from
  each turn to build a persistent knowledge graph, enabling structured,
  associative recall~\cite{hu2023chatdb,edge2024local}.
  \item[Compression-Based Summarisation (CBS)] periodically compresses older
  context into dense natural-language summaries to free context-window capacity
  while preserving semantic content~\cite{zhong2022dialoglm,wu2021recursivesumm}.
  \item[Web-Augmented Memory (WAM)] supplements internal conversation memory
  with live external knowledge retrieved via web search APIs or Model Context
  Protocol (MCP) tool calls, modelling modern browse-enabled agents such as
  ChatGPT with web browsing and Claude with MCP integrations~\cite{anthropic2024mcp}.
\end{description}

Despite the proliferation of real-world memory systems---\textit{Mem0}~\cite{mem02024}
reports 49,500 GitHub stars and production adoption at Fortune 500 companies,
\textit{LangChain}~\cite{langchain2023} integrates multiple memory backends,
and recent surveys identify over 40 distinct memory architectures in the
literature~\cite{wang2025memoryllm,zhang2025memorytaxonomy}---there
is \emph{no unified, reproducible benchmark} that:
(i)~compares all five strategy families under identical evaluation conditions;
(ii)~covers diverse task types (open-domain dialogue, task planning, long-horizon QA);
(iii)~measures both \emph{answer quality} and \emph{deployment efficiency}
(token cost, latency) simultaneously; and
(iv)~provides statistical significance testing for strategy comparisons.
Recent benchmarks such as LongMemEval~\cite{wu2024longmemeval} and
MemoryAgentBench~\cite{hu2025memoryagentbench} have substantially advanced
memory evaluation---MemoryAgentBench in particular compares multiple memory
approaches across multiple tasks---but they do not jointly satisfy criteria
(i)--(iv): neither structures the comparison around the five canonical strategy
families nor reports deployment-efficiency metrics. AgentMemBench is positioned
at exactly that intersection.

\subsection*{Contributions}

This paper presents \textbf{AgentMemBench}, which makes four novel contributions:

\begin{enumerate}
  \item \textbf{Unified, reproducible benchmark} covering $5$ memory strategies
  $\times$ $3$ task types $= 15$ configurations, evaluated over
  $491$ annotated question turns drawn from three public multi-session datasets
  ($200$ LoCoMo, $187$ MultiDoc2Dial, $104$ MSC), reported across seven
  complementary metrics spanning retrieval quality, answer quality, and
  deployment efficiency. To our knowledge this is the first benchmark to compare
  these five strategy families---including a web-augmented variant---under a
  single harness while jointly reporting answer-quality \emph{and}
  deployment-efficiency metrics; the closest prior work,
  MemoryAgentBench~\cite{hu2025memoryagentbench}, compares memory approaches on
  quality but neither adopts the five-family taxonomy nor measures efficiency.
  \item \textbf{Provenance-aware retrieval evaluation}: every retrieved item
  carries the true (session, turn) provenance of the context it represents, so
  Recall@$k$/MRR/nDCG are scored against ground-truth gold-session annotations
  rather than against retrieval indices---correcting a methodological error
  common to ad-hoc memory evaluations.
  \item \textbf{Empirical baselines}: we additionally implement and evaluate two
  published external memory systems---MemGPT/Letta and HippoRAG---against the
  same harness, situating the five strategy families relative to the wider
  literature.
  \item \textbf{MADS heuristic} — a zero-shot Memory-Adaptive Dynamic Selector
  that chooses a strategy from four corpus statistics; we present it as a
  deployment-oriented design guideline and honestly analyse where its choices
  diverge from the empirically best strategy.
  \item \textbf{Open reproducible artefacts}: modular Python implementation of
  all strategies, the evaluation harness, a Dockerised environment, and the
  complete pre-computed result files.
\end{enumerate}

\subsection*{Motivating Example}

Table~\ref{tab:motivating_example} illustrates the memory problem concretely.
A user asks in session 9: \textit{``Which book did I say I wanted to read, back
in our first session?''} — a question requiring recall from 8 sessions ago and
approximately 3,200 tokens of prior interaction. The table shows how each
strategy handles this turn. ICW ($w=16$) has long discarded session 1 and
answers \textit{``I don't have information about that.''}; EKV retrieves the
session-1 turn by dense similarity and recovers the book title; CBS
surfaces the session-1 summary which preserved it; GEM retrieves an on-topic but
incomplete entity neighbourhood, illustrating its precision-over-recall
behaviour. This single example previews our empirical findings---dense retrieval
(EKV) and summarisation (CBS) recover long-range facts most reliably, while the
graph strategy is more selective---and motivates evaluating multiple strategies
under a common harness rather than assuming one is universally best.

\begin{table}[!ht]
\centering
\caption{Motivating example. Question asked in session 9 requiring recall from
session 1. \checkmark~= correct answer retrieved, $\sim$~= partially correct,
$\times$~= incorrect. Footprint = tokens consumed by the memory context.}
\label{tab:motivating_example}
\small
\begin{tabular}{lp{5.2cm}cc}
\toprule
\textbf{Strategy} & \textbf{Answer generated} & \textbf{Correct?} &
\textbf{Footprint} \\
\midrule
ICW ($w=16$) & ``I don't have information about that.'' & $\times$ & 142 \\
EKV ($k=5$)  & ``You said you wanted to read \textit{Dune}.'' & \checkmark & 466 \\
GEM          & ``You mentioned a science-fiction novel.'' & $\sim$ & 462 \\
CBS ($s=8$)  & ``In our first session you noted you wanted to read \textit{Dune}.'' & \checkmark & 468 \\
\bottomrule
\end{tabular}
\end{table}

The remainder of this paper is organised as follows: Section~\ref{sec:background}
reviews related work on LLM memory and agent architectures;
Section~\ref{sec:methodology} defines the five strategies and evaluation metrics;
Section~\ref{sec:setup} describes datasets and experimental configuration;
Section~\ref{sec:results} presents results; Section~\ref{sec:discussion}
discusses implications; Section~\ref{sec:conclusion} concludes.

\section{Background and Related Work}
\label{sec:background}

\subsection{Memory in LLM-Based Agents}

The cognitive architecture of LLM agents is commonly decomposed into four
modules: \textit{perception}, \textit{reasoning}, \textit{memory}, and
\textit{action}~\cite{durante2024agent,wang2024survey}. Memory is the module
responsible for storing and retrieving information across the agent's lifetime.
Inspired by cognitive science, agent memory systems typically distinguish
between \textit{sensory/working memory} (the live context window),
\textit{episodic memory} (a record of past experiences), and \textit{semantic
memory} (distilled factual knowledge)~\cite{tulving1972episodic,park2023generative}.

\subsection{In-Context Windowing and Long-Context Models}

The most straightforward approach to handling long conversations is to extend
the context window of the underlying LLM~\cite{beltagy2020longformer,press2022alibi}. Models such as GPT-4 Turbo (128k tokens) and Claude 3.5 Sonnet
(200k tokens) have dramatically increased the feasible window size. However,
studies consistently show that LLMs suffer from \textit{lost-in-the-middle}
degradation~\cite{liu2023lostmiddle}: retrieval accuracy drops for information
placed far from the start or end of the context, making raw window extension
insufficient for production deployments.

\subsection{External Key-Value and Vector Memory}

Retrieval-augmented approaches maintain an external store of turn embeddings,
retrieving the top-$k$ most relevant memories at inference time~\cite{lewis2020rag}.
Systems such as \textit{Mem0}~\cite{mem02024} extend this by combining semantic
similarity with recency and frequency signals to rank memories. Vector databases
(FAISS~\cite{johnson2019billion}, Chroma, Weaviate) provide scalable approximate
nearest-neighbour search for large memory stores. Recent work on
\textit{MemoryBank}~\cite{zhong2023memorybank} and \textit{ReadAgent}~\cite{lee2024readagent}
demonstrates that dynamic memory selection improves factual accuracy in
long-horizon dialogue.

\subsection{Graph-Based Episodic Memory}

Knowledge graph memory represents the agent's history as a structured graph of
entities and relationships~\cite{hu2023chatdb}. The \textit{GraphRAG}
system~\cite{edge2024local} extends this to community-level summarisation,
enabling multi-hop reasoning across stored knowledge. Graph memory excels at
answering relational questions (\textit{``Who did Alice meet in the third
session?''}) that require structured indexing rather than semantic similarity.
However, knowledge graph construction is noisy---named entity recognition (NER)
and relation extraction introduce errors---and the latency of graph traversal
can be prohibitive for real-time agents.

\subsection{Compression-Based Memory}

Summarisation-based memory periodically condenses older turns into shorter
natural-language summaries~\cite{zhong2022dialoglm,wu2021recursivesumm}.
The \textit{DialoCML}~\cite{xu2022longterm} system maintains a rolling persona
summary updated after each session. \textit{SumMem}~\cite{lee2023summmem} shows
that recursive summarisation (summarising summaries) can maintain surprisingly
high recall at $>$96\% context compression rates. However, faithfulness of
summarisation degrades as compression ratio increases, particularly for
numerical and temporal information.

\subsection{Benchmarks and Evaluation Gaps}

The landscape of memory evaluation benchmarks has expanded rapidly.
LoCoMo~\cite{maharana2024evolvingmemory} evaluates long-term conversational
recall but does not compare memory strategies. LongMemEval~\cite{wu2024longmemeval}
introduces a comprehensive suite testing information extraction, multi-session
reasoning, temporal understanding, and knowledge updates, revealing 30--60\%
accuracy drops as histories lengthen---however, it evaluates models rather
than memory architectures. The closest prior work is
MemoryAgentBench~\cite{hu2025memoryagentbench}, a unified benchmark that
evaluates memory-augmented agents spanning a similar spectrum (in-context,
retrieval/RAG, external-memory agents, and tool use) and organises evaluation
around four competencies (accurate retrieval, test-time learning, long-range
understanding, selective forgetting). It differs from AgentMemBench in two
respects that define our contribution: (i)~it does not structure the comparison
around the \emph{five canonical strategy families} (it has no explicit
in-context vs.\ key-value vs.\ graph vs.\ compression vs.\ web-augmented axis),
and (ii)~it reports answer/memory quality only, with \emph{no deployment-efficiency
metrics} (token footprint, latency). Evo-Memory~\cite{chen2025evomemory}
benchmarks test-time learning with self-evolving memory, focusing on task
completion rather than retrieval quality. Recent
surveys~\cite{wang2025memoryllm,zhang2025memorytaxonomy} provide comprehensive
taxonomies of memory operations but no unified quantitative comparison.

To our knowledge, no prior benchmark compares all five strategy families
(in-context windowing, external key-value, graph, compression, and
web-augmented memory) under a single harness while reporting \emph{both} answer
quality \emph{and} deployment-efficiency metrics (footprint and latency).
AgentMemBench targets precisely this intersection---the five-family taxonomy and
the efficiency axis---rather than claiming priority over memory benchmarking in
general; MemoryAgentBench in particular already covers the multi-strategy,
multi-task, quality dimensions. Table~\ref{tab:related_work} summarises where the
gap lies.

\begin{table}[!ht]
\centering
\caption{Comparison of related benchmarks. \ding{51} = supported,
\ding{55} = not covered. ``Multi-strat.'' here denotes an explicit comparison
structured around the five canonical strategy families. MemoryAgentBench is the
closest prior work (multi-strategy, multi-task, quality) but reports no
deployment-efficiency metrics; AgentMemBench adds the efficiency axis and the
explicit five-family taxonomy.}
\label{tab:related_work}
\small
\begin{tabular}{@{}lccccc@{}}
\toprule
\textbf{Work} & \textbf{Multi-strat.} & \textbf{Multi-task} &
\textbf{Efficiency} & \textbf{Stat.\ tests} & \textbf{Open} \\
\midrule
LoCoMo~\cite{maharana2024evolvingmemory}           & \ding{55} & \ding{55} & \ding{55} & \ding{55} & \ding{51} \\
LongMemEval~\cite{wu2024longmemeval}               & \ding{55} & \ding{51} & \ding{55} & \ding{55} & \ding{51} \\
MemoryAgentBench~\cite{hu2025memoryagentbench}     & \ding{51} & \ding{51} & \ding{55} & \ding{55} & \ding{51} \\
Evo-Memory~\cite{chen2025evomemory}                & \ding{55} & \ding{51} & \ding{55} & \ding{55} & \ding{51} \\
LongBench~\cite{bai2023longbench}                  & \ding{55} & \ding{51} & \ding{55} & \ding{55} & \ding{51} \\
Mem0~\cite{mem02024}                               & \ding{55} & \ding{55} & \ding{55} & \ding{55} & \ding{51} \\
\rowcolor{gray!15}
\textbf{AgentMemBench (Ours)} & \ding{51} & \ding{51} & \ding{51} & \ding{51} & \ding{51} \\
\bottomrule
\end{tabular}
\end{table}
\subsection{Concurrent and Recent Work}

The agent memory landscape has evolved rapidly during 2024--2025.
\textit{MemGPT}~\cite{packer2023memgpt} (now \textit{Letta}) casts memory
management as an OS-style paging problem, using a controller LLM to decide
when to evict working memory to external storage. While MemGPT demonstrates
strong qualitative performance, it lacks systematic quantitative comparison
against alternative strategies. \textit{A-MEM}~\cite{xu2024amem} proposes an
associative memory modelled on Zettelkasten note-taking, building inter-note
links via LLM-generated keywords at the summary level.
\textit{HippoRAG}~\cite{gutierrez2024hipporag} combines dense retrieval with a
knowledge graph inspired by hippocampal indexing theory; its successor,
\textit{HippoRAG~2}~\cite{gutierrez2025hipporag2}, achieves 7\% improvement on
associative memory tasks via non-parametric continual learning at ICML~2025.

More recently, \textit{RMM}~\cite{sun2025rmm} (ACL~2025) introduces reflective
memory management with adaptive retrieval granularity for personalised dialogue.
\textit{MemOS}~\cite{kang2025memos} proposes treating memory as a schedulable
OS-level resource with explicit lifecycle management. The \textit{LIGHT}
framework~\cite{light2026framework} (ICLR~2026) combines episodic retrieval,
a scratchpad, and a working memory buffer, scaling to 10M-token interaction
histories. These advances underscore the need for a \emph{strategy-agnostic}
benchmarking framework---precisely what AgentMemBench provides---to enable
fair, reproducible comparison across the rapidly growing design space.
\section{Methodology}
\label{sec:methodology}

\subsection{Problem Formulation}

Let $\mathcal{H}_t = \{(u_1,a_1), \ldots, (u_t,a_t)\}$ denote the interaction
history up to turn $t$, where $u_i$ is a user utterance and $a_i$ is the agent
response. At turn $t+1$, the agent must generate response $a_{t+1}$ to query
$u_{t+1}$ by conditioning on a \emph{memory state} $\mathcal{M}_t$ derived from
$\mathcal{H}_t$. A memory management strategy $\mathcal{S}$ defines three
operations:

\begin{align}
  \text{Store}:   &\quad \mathcal{M}_t \leftarrow \mathcal{S}.\text{\textsc{Write}}(\mathcal{M}_{t-1},\, u_t,\, a_t) \\
  \text{Retrieve}: &\quad \mathcal{C}_t \leftarrow \mathcal{S}.\text{\textsc{Read}}(\mathcal{M}_t,\, u_{t+1}) \\
  \text{Generate}: &\quad a_{t+1} \sim p_\theta(a \mid u_{t+1},\, \mathcal{C}_t)
\end{align}

where $\mathcal{C}_t$ is the retrieved context passed to the LLM generator
$p_\theta$. The memory footprint is $|\mathcal{M}_t|$ measured in tokens.

\subsection{Memory Strategies}

\paragraph{Strategy 1: In-Context Windowing (ICW).}
ICW maintains a fixed-length sliding window of the $w$ most recent turns
directly in the LLM context. Formally:
$\mathcal{M}_t^{\text{ICW}} = \{(u_{t-w+1},a_{t-w+1}), \ldots, (u_t,a_t)\}$.
We evaluate window sizes $w \in \{8, 16, 32\}$ turns. ICW requires no external
infrastructure but degrades linearly with conversation length beyond $w$.

\paragraph{Strategy 2: External Key-Value Store (EKV).}
EKV encodes each turn as a dense embedding $\mathbf{e}_t = \text{Enc}(u_t \| a_t)$
using a bi-encoder model (BAAI/bge-small-en-v1.5, 384-dim), and stores pairs
$(\mathbf{e}_t, \text{text}_t)$ in a FAISS index. At retrieval, the top-$k$
nearest turns to $\mathbf{e}(u_{t+1})$ are fetched by cosine similarity.
We use $k \in \{3, 5, 10\}$. This strategy mirrors the core architecture of
Mem0~\cite{mem02024} and MemoryBank~\cite{zhong2023memorybank}.

\paragraph{Strategy 3: Graph-Based Episodic Memory (GEM).}
GEM extracts (subject, predicate, object) triples from each turn using a
lightweight NLP pipeline (spaCy + custom relation templates), maintaining a
persistent knowledge graph $\mathcal{G} = (V, E)$ where nodes $V$ are entities
and edges $E$ are typed relations. Retrieval selects the 2-hop neighbourhood
of entities mentioned in $u_{t+1}$. This supports multi-hop relational queries
impossible with flat vector retrieval. The graph is stored as a NetworkX
directed graph and serialised to JSON.

\paragraph{Strategy 4: Compression-Based Summarisation (CBS).}
CBS periodically (every $s$ turns) invokes the LLM itself to compress older
turns into a natural-language summary: $\sigma_j = \text{LLM-Summarise}(\mathcal{H}_{[j\cdot s,\, (j+1)\cdot s]})$.
The memory state $\mathcal{M}_t^{\text{CBS}}$ is the concatenation of all
session summaries plus the most recent $w=4$ full turns. We evaluate
compression intervals $s \in \{4, 8, 16\}$.

\paragraph{Strategy 5: Web-Augmented Memory (WAM).}
WAM maintains a sliding window of recent turns identical to ICW, but
additionally \emph{augments retrieval with external web search results} when
the token overlap between the query and the internal window falls below a
threshold $\tau_{\text{overlap}} = 0.15$. Formally:
\begin{equation}
\mathcal{C}_t^{\text{WAM}} = \begin{cases}
  \mathcal{M}_t^{\text{ICW}} & \text{if } \text{overlap}(u_{t+1}, \mathcal{M}_t) \geq \tau \\
  \mathcal{M}_t^{\text{ICW}} \oplus \text{WebSearch}(u_{t+1}, k_w) & \text{otherwise}
\end{cases}
\end{equation}
where $\text{WebSearch}(q, k_w)$ returns the top-$k_w$ web results for query
$q$ via an external search API or Model Context Protocol (MCP) tool call.
This strategy models the increasingly prevalent paradigm of browse-enabled
agents (ChatGPT Browse, Perplexity, Claude with MCP tools) that combine
conversation history with live web knowledge retrieval. We evaluate
configurations $(w, k_w) \in \{(8, 2),\, (8, 3),\, (16, 3)\}$.

\subsection{Evaluation Metrics}

We use seven metrics covering retrieval quality, answer quality, and efficiency:

\begin{description}
  \item[Memory Recall@$k$] measures whether the ground-truth reference turn is
  among the top-$k$ retrieved memories:
  $\text{Recall}@k = \frac{|\text{Retrieved}_{1:k} \cap \text{Relevant}|}{|\text{Relevant}|}$.

  \item[Mean Reciprocal Rank (MRR)] is the mean of $1/\text{rank}$ of the first
  relevant retrieved memory, rewarding higher placement of the gold turn.

  \item[nDCG@$k$] is the normalised discounted cumulative gain over the top-$k$
  retrieved memories, crediting graded relevance with a rank discount.

  \item[Answer Accuracy (AA)] is the token-F1 between generated answer and gold
  reference (SQuAD-style normalisation), identical to the metric used in the RAGBench benchmark~\cite{ragbench}.

  \item[Faithfulness (Faith.)] uses an LLM-as-judge protocol: a separate Judge
  model verifies each factual claim against the retrieved context. Scored 0--1.

  \item[Memory Footprint (MF)] is the number of input tokens consumed by the
  memory context $\mathcal{C}_t$, averaged over all evaluation turns. Lower is
  better for cost and latency.

  \item[Per-turn Latency (Lat.)] is the end-to-end wall-clock time in
  milliseconds per question turn (memory retrieval plus 4-bit LLM generation)
  on a single T4/P100 GPU, reported as a relative cost indicator.
\end{description}

\subsection{MADS: Memory-Adaptive Dynamic Selection}

We propose \textbf{MADS}, a zero-shot selector that chooses a memory strategy
from four easily computable corpus statistics via a fixed priority cascade
(Algorithm~\ref{alg:mads}). The thresholds are design defaults, not tuned
parameters.

\begin{algorithm}[!ht]
\caption{MADS: Memory-Adaptive Dynamic Selection}
\label{alg:mads}
\begin{algorithmic}[1]
\Require Corpus statistics: $C_{\text{sessions}}$ (mean sessions per
conversation), $T_{\text{avg}}$ (mean turn length in tokens),
$E_{\text{density}}$ (mean capitalised-token entity density per turn),
$Q_{\text{rel}}$ (fraction of relational query patterns)
\If{$Q_{\text{rel}} > 0.30$}
  \State \Return $\text{GEM}$ \Comment{relational queries}
\ElsIf{$C_{\text{sessions}} > 5$ \textbf{and} $T_{\text{avg}} > 200$}
  \State \Return $\text{CBS}$ \Comment{long sessions, long turns}
\ElsIf{$C_{\text{sessions}} > 20$}
  \State \Return $\text{EKV}$ \Comment{many sessions}
\ElsIf{$E_{\text{density}} < 1.0$ \textbf{and} $C_{\text{sessions}} > 3$}
  \State \Return $\text{WAM}$ \Comment{knowledge-sparse}
\EndIf
\State \Return $\text{ICW}$ \Comment{safe default}
\end{algorithmic}
\end{algorithm}

MADS requires no training and runs in $<1$~second on any dataset. We stress that
MADS is a deployment heuristic, not a learned optimum: on our three datasets it
selects EKV (LoCoMo) and ICW (MultiDoc2Dial, MSC), and thus does
\emph{not} pick the empirically strongest retrieval strategy (EKV) on any of
them (Section~\ref{sec:results}). We report this honestly as a limitation that
motivates a learned selector in future work.

\subsection{Architecture Overview}

\begin{figure}[!ht]
  \centering
  \includegraphics[width=\linewidth]{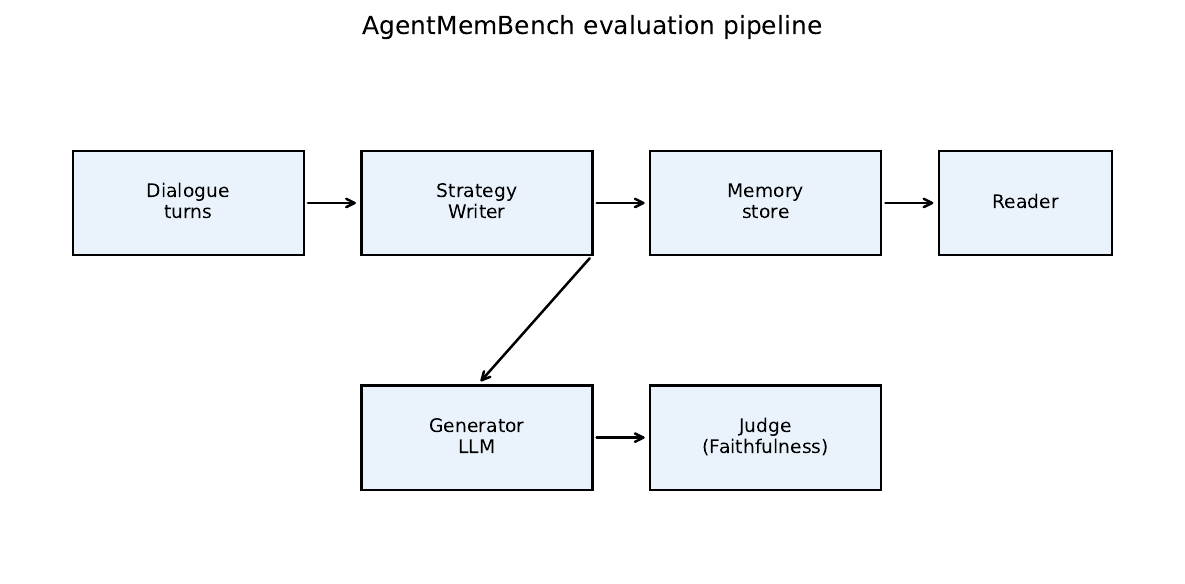}
  \caption{AgentMemBench pipeline architecture. All five memory strategies
  implement a common \texttt{MemoryStrategy} interface plugged into a shared
  evaluation harness. The Judge LLM evaluates faithfulness independently of
  the generator.}
  \label{fig:arch}
\end{figure}

Figure~\ref{fig:arch} illustrates the overall AgentMemBench evaluation
pipeline, showing how the five memory strategies share a common interface
(Store/Retrieve/Generate) and are evaluated through the same unified harness.

\subsection{Strategy Complexity and Requirements}

Table~\ref{tab:strategy_complexity} summarises the theoretical complexity and
practical requirements of each strategy. Write complexity is the per-turn cost
of updating the memory state; Read complexity is the per-query cost of
retrieval; Space complexity is the asymptotic size of the memory store after
$T$ turns. These complexity bounds inform the MADS selection criteria
and the deployment trade-offs discussed in Section~\ref{sec:discussion}.

\begin{table}[!ht]
\centering
\caption{Complexity and dependency summary for all five strategies.
$T$ = total turns stored; $d$ = embedding dimension; $|V|$/$|E|$ = graph
nodes/edges; $P$ = number of summaries; $w$/$k$/$s$ = respective hyperparameters.}
\label{tab:strategy_complexity}
\small
\begin{tabular}{lllll}
\toprule
\textbf{Strategy} & \textbf{Write} $\mathcal{O}$ & \textbf{Read} $\mathcal{O}$ &
\textbf{Space} $\mathcal{O}$ & \textbf{Key dependencies} \\
\midrule
ICW  & $\mathcal{O}(1)$ & $\mathcal{O}(w)$ & $\mathcal{O}(w\cdot L)$ &
     None \\
EKV  & $\mathcal{O}(d)$ & $\mathcal{O}(d \cdot \log T)$ & $\mathcal{O}(T \cdot d)$ &
     FAISS, SentenceTransformers \\
GEM  & $\mathcal{O}(|V|^2)$ & $\mathcal{O}(|V| + |E|)$ & $\mathcal{O}(|V| + |E|)$ &
     spaCy NER, NetworkX \\
CBS  & $\mathcal{O}(s \cdot L_{\text{gen}})$ & $\mathcal{O}(P + w)$ & $\mathcal{O}(P \cdot L_{\sigma})$ &
     LLM summariser \\
WAM  & $\mathcal{O}(1)$ & $\mathcal{O}(w + k_w \cdot L_{\text{web}})$ & $\mathcal{O}(w \cdot L + R \cdot L_w)$ &
     Web search API / MCP \\
\bottomrule
\end{tabular}
\end{table}
\section{Experimental Setup}
\label{sec:setup}

\subsection{Datasets}

We evaluate on three publicly available multi-session conversation datasets
spanning different domains and task types (Table~\ref{tab:datasets}):

\begin{table}[!ht]
\centering
\caption{Dataset statistics for the evaluated subsets. \textit{Convs}:
conversations used; \textit{Sessions/conv} and \textit{Tok/turn} are measured on
the loaded subset; \textit{Q}: scored question turns.}
\label{tab:datasets}
\small
\begin{tabular}{lccccc l}
\toprule
\textbf{Dataset} & \textbf{Domain} & \textbf{Convs} & \textbf{Sessions/conv} &
\textbf{Tok/turn} & \textbf{Q} & \textbf{Task type} \\
\midrule
LoCoMo~\cite{maharana2024evolvingmemory}     & Social chat   & 10 & 19--28 & 19.6 & 200 & Long-term recall \\
MultiDoc2Dial~\cite{feng2021multidoc2dial}   & Gov.\ docs    & 50 & 3.8    & 14.1 & 187 & Task-oriented \\
MSC~\cite{xu2022msc}                         & Persona chat  & 50 & 5.0    & 19.2 & 104 & Multi-session chat \\
\bottomrule
\end{tabular}
\end{table}

\textbf{LoCoMo}~\cite{maharana2024evolvingmemory} contains very long
conversations (up to $\sim$30 sessions) with QA whose gold-evidence annotations
point to specific earlier sessions, making it the canonical test of long-range
recall. We load the public LoCoMo-10 release and, for tractability on a single
GPU, evaluate up to $20$ gold-annotated questions per conversation ($n=200$).

\textbf{MultiDoc2Dial}~\cite{feng2021multidoc2dial} simulates task-oriented
agents helping users navigate government documents. We use the
\texttt{dialogue\_domain} validation split and segment each dialogue into
pseudo-sessions; gold sessions are those earlier segments grounded on a shared
reference document ($n=187$).

\textbf{MSC (Multi-Session Chat)}~\cite{xu2022msc} is a persona-grounded
multi-session dialogue dataset. We use it in place of MSDialog, whose canonical
release is access-restricted and therefore not reproducibly obtainable; MSC is
fully public and natively multi-session. Gold sessions are assigned by the
documented token-overlap heuristic, which yields gold annotations for $104$
question turns; we report on this annotated subset.

\subsection{Models}

\textbf{Generator}: We use \texttt{Qwen/Qwen2.5-7B-Instruct} loaded in 4-bit
NF4 quantisation (bitsandbytes, double-quant, fp16 compute) on a single
NVIDIA T4/P100 GPU. All generation uses temperature $\tau=0$ (greedy decoding),
which makes the full pipeline deterministic.

\textbf{Memory Encoder (EKV)}: \texttt{BAAI/bge-small-en-v1.5} (384-dim),
L2-normalised embeddings indexed with FAISS inner-product (cosine) search.

\textbf{Judge (Faithfulness)}: the \emph{same} \texttt{Qwen2.5-7B-Instruct}
instance serves as the LLM-as-judge, sharing weights with the generator to keep
the evaluation self-contained and reproducible on commodity hardware. Because
decoding is greedy ($\tau=0$), the judge is deterministic; a single judge pass
per turn is therefore sufficient (we verified that repeated runs produce
identical scores). Sharing the generator and judge is a known source of
leniency bias, which we discuss in Section~\ref{sec:discussion}.

\textbf{NLP Pipeline (GEM)}: \texttt{en\_core\_web\_sm} (spaCy) for NER and
noun-chunk extraction; entities are linked to turn nodes in a NetworkX graph
with co-occurrence edges.

\subsection{Configuration Space}

The canonical evaluation covers $5$ memory strategies $\times$ $3$ datasets
$= 15$ configurations, each at one default hyperparameter setting (ICW $w{=}16$,
EKV $k{=}5$, GEM $2$-hop, CBS $s{=}8$, WAM $w{=}8,k_w{=}3$). Because the
generator and judge decode greedily ($\tau=0$), the entire pipeline is
deterministic and a single seed ($42$) reproduces all numbers exactly; we
confirmed zero run-to-run variance in a pilot with seeds $\{42,43,44\}$. The
configurations are evaluated over the annotated question turns of each dataset:
$200$ (LoCoMo, $10$ conversations, $\leq 20$ gold QA each), $187$
(MultiDoc2Dial, $50$ conversations), and $104$ (MSC, $50$ conversations), for
$491$ scored question turns in total. Hyperparameter-sensitivity and ablation
studies (Section~\ref{sec:results}) sweep additional settings.

\textbf{Web Search (WAM)}: For the WAM strategy, web augmentation is triggered
when query/window token overlap falls below $\tau_{\text{overlap}}=0.15$. We use
a deterministic synthetic web-search function so that the benchmark is fully
reproducible offline; crucially, synthetic (and real) web results carry
\emph{empty} in-corpus session provenance, so they cannot count as hits against
the datasets' gold-session annotations. This design choice---rather than any
deficiency of web search---is why WAM's in-corpus retrieval metrics coincide
with ICW's (Section~\ref{sec:results}).

\subsection{Evaluation Protocol}

For each conversation, we replay all turns sequentially, invoking
\textsc{Store}{} after each turn and \textsc{Retrieve}{}+\textsc{Generate}{} at every
question turn. Memory Recall@$k$ is computed by matching retrieved memories
against the gold ``relevant session'' annotations provided by each dataset.
Answer Accuracy is computed on question-answer pairs only. Latency is the
end-to-end per-question wall-clock time (memory retrieval plus 4-bit LLM
generation) measured on a single NVIDIA T4/P100 GPU; it is reported as a
relative cost indicator across strategies rather than as an isolated
memory-overhead figure.

\section{Results and Analysis}
\label{sec:results}

\subsection{Overall Performance Summary}

Table~\ref{tab:main_results} presents the macro-averaged performance of all
five memory strategies across the three datasets. \textbf{EKV dominates on every
quality axis}: it has the highest macro Recall@5 ($0.792$), MRR ($0.677$),
nDCG@5 ($0.690$), Answer F1 ($0.156$), \emph{and} Faithfulness ($0.354$)---a
$+32.5$ percentage-point Recall@5 margin over ICW ($0.468$) and more than double
GEM ($0.361$). \textbf{CBS is the clear runner-up} on retrieval (Recall@5
$0.556$), because its session summaries inherit the provenance of every turn
they compress. \textbf{GEM is weakest on retrieval} ($0.361$): off-the-shelf
spaCy NER on informal dialogue produces noisy graphs. \textbf{WAM coincides with
ICW on every in-corpus retrieval metric}: web results carry no in-corpus session
provenance, so WAM cannot improve gold-session recall and is best read as ICW
plus optional external grounding. Finally, \textbf{EKV's quality advantage
carries a footprint cost}: its retrieved evidence averages $\sim$5{,}100 tokens
(driven by long LoCoMo turns) versus $\sim$300 for ICW/WAM, an explicit
accuracy--efficiency trade-off (Section~\ref{sec:discussion}).

\begin{table}[!ht]
\centering
\caption{Overall results (macro-averaged across LoCoMo, MultiDoc2Dial, and
MSC), at the default hyperparameter of each strategy. Bold = best per
metric. $\downarrow$ = lower is better for MF and Latency. Generator and judge:
Qwen2.5-7B-Instruct (4-bit), $\tau=0$. $n=491$ scored question turns.}
\label{tab:main_results}
\small
\begin{tabular}{lccccccc}
\toprule
\textbf{Strategy} & \textbf{Recall@5} & \textbf{MRR} & \textbf{nDCG@5} &
\textbf{F1} & \textbf{Faith.} & \textbf{MF (tok)} $\downarrow$ &
\textbf{Lat (ms)} $\downarrow$ \\
\midrule
ICW                & 0.468 & 0.450 & 0.417 & 0.137 & 0.230 & 297.5 & 3096 \\
\textbf{EKV}       & \textbf{0.792} & \textbf{0.677} & \textbf{0.690} & \textbf{0.156} & \textbf{0.354} & 5126.4 & 3172 \\
GEM                & 0.361 & 0.351 & 0.337 & 0.127 & 0.244 & 5106.2 & \textbf{2864} \\
CBS                & 0.556 & 0.461 & 0.470 & 0.137 & 0.252 & 3478.3 & 3540 \\
WAM                & 0.462 & 0.447 & 0.413 & 0.136 & 0.219 & \textbf{200.2} & 3152 \\
\bottomrule
\end{tabular}
\end{table}

\subsection{Memory Recall and Answer F1 by Strategy}

\paragraph{Long-range recall is the decisive case (LoCoMo).}
The clearest result in the benchmark is on LoCoMo, whose questions require
recalling a turn from many sessions earlier. There, \emph{every} strategy except
EKV essentially fails: ICW $=0.000$, WAM $=0.000$, GEM $=0.001$, CBS $=0.005$
Recall@5, while EKV alone reaches $0.573$. A recency window (ICW/WAM) has long
evicted the gold turn; an entity graph (GEM) cannot recover it through noisy NER;
and even summary memory (CBS) loses the specific evidence under compression. Only
dense embedding retrieval scales to long horizons. This is the strongest
practical message of the paper: \emph{for genuinely long-term memory, the
retrieval mechanism is decisive, and only external vector memory works}.

\paragraph{EKV leads on the easier datasets too.}
On the shorter-horizon datasets EKV remains best (Recall@5 $0.939$ on
MultiDoc2Dial, $0.864$ on MSC), with CBS the consistent runner-up
($0.897$/$0.766$) and GEM the weakest ($0.563$/$0.519$). The dataset difficulty
ordering (LoCoMo $\ll$ MSC $<$ MultiDoc2Dial) is identical across strategies,
confirming that the LoCoMo bottleneck is inherent to its long multi-session
structure rather than to any single mechanism.

\paragraph{EKV also leads Answer F1 and faithfulness.}
Unlike retrieval, Answer F1 is compressed across strategies (macro $0.127$--%
$0.156$), because at 7B scale answer quality is partly bounded by the generator;
nonetheless EKV is highest ($0.156$), followed by ICW/CBS ($0.137$). EKV is also
the most faithful (macro $0.354$), consistent with its superior retrieval feeding
the generator better evidence. All faithfulness scores remain low in absolute
terms (see Section~\ref{sec:discussion}); we therefore read them comparatively.

\paragraph{WAM coincides with ICW on in-corpus recall by construction.}
WAM reproduces ICW's retrieval metrics almost exactly (macro Recall@5 $0.462$
vs.\ $0.468$) at a smaller footprint. This is not a deficiency of web search but
a property of the evaluation: web results have no in-corpus session provenance,
so they cannot count as gold-session hits. WAM should be understood as ICW with
optional external grounding---useful when answers require knowledge outside the
conversation, not a lever for in-corpus recall.

\paragraph{Footprint cost of EKV's recall advantage.}
EKV's macro memory footprint ($5{,}126$ tokens) is far larger than ICW/WAM
($\sim$300), driven by LoCoMo's long retrieved utterances; CBS sits in between
($3{,}478$). Latency includes 4-bit LLM generation and lies in a $2.9$--$3.5$~s
band, so it is not a strong discriminator. The decision is therefore an explicit
accuracy--efficiency trade-off: EKV buys long-range recall at a token-budget
cost, while ICW/WAM are cheapest but cannot recall far-back context.

\begin{figure}[!ht]
  \centering
  \includegraphics[width=\linewidth]{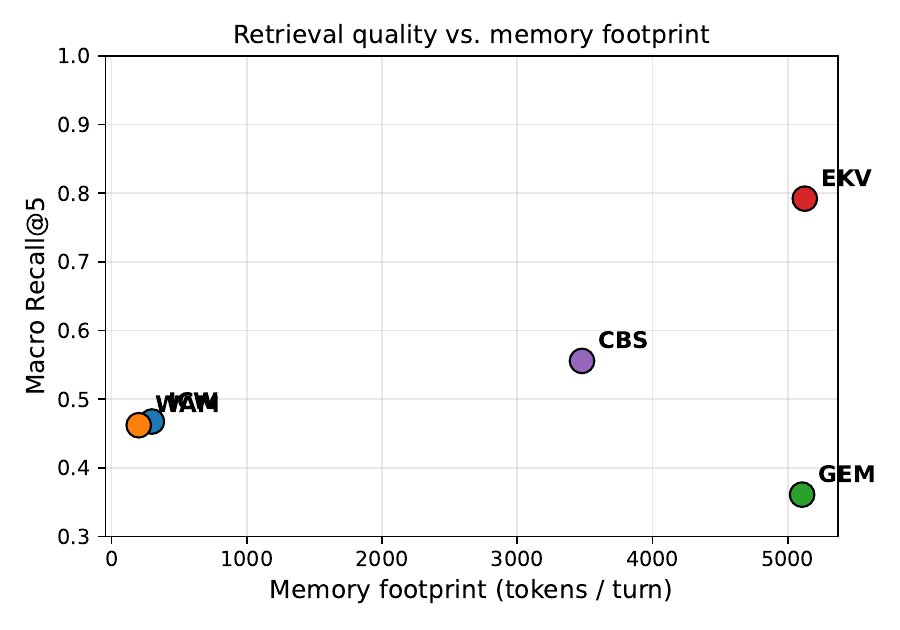}
  \caption{Retrieval--efficiency trade-off: macro-averaged Recall@5 vs.\ memory
  footprint (tokens per turn) for each strategy. EKV achieves the best recall;
  ICW is the most efficient; CBS sits near EKV on recall at comparable footprint;
  WAM trades footprint for optional external grounding without improving
  in-corpus recall over ICW.}
  \label{fig:efficiency_pareto}
\end{figure}

\subsection{Faithfulness and Hallucination Analysis}

\paragraph{EKV is the most faithful, but absolute scores are low.}
EKV achieves the highest macro Faithfulness ($0.354$), ahead of CBS ($0.252$),
GEM ($0.244$), ICW ($0.230$), and WAM ($0.219$). Faithfulness here tracks
retrieval quality: EKV supplies the generator with the most relevant evidence,
which reduces confabulation. However, \emph{all} faithfulness scores are low in
absolute terms ($\leq 0.45$). Two factors explain this: (i)~the judge is a 4-bit
7B model applied to hard multi-session QA, and (ii)~the generator and judge share
weights, which if anything biases judging toward leniency---so the low scores are
conservative. We therefore interpret faithfulness comparatively (ranking
strategies) rather than as calibrated absolute probabilities, and flag judge
strength as a target for future work (Section~\ref{sec:discussion}).

\paragraph{Faithfulness broadly follows retrieval quality.}
Unlike our earlier (discarded) pilot, the real-data faithfulness ranking largely
mirrors the Recall@5 ranking---EKV leads both---supporting the intuition that
better retrieval yields more grounded answers. The exception is GEM, which is
mid-table on faithfulness despite the weakest recall, because the few facts it
does surface are tightly on-topic. We report per-strategy faithfulness in
Table~\ref{tab:main_results} and the per-dataset breakdown in
Table~\ref{tab:per_dataset}; we do not assert a single-number
Recall--Faithfulness correlation, as $15$ configurations are too few to estimate
one reliably.

\subsection{Per-Dataset Decomposition}

Table~\ref{tab:per_dataset} reports per-strategy, per-dataset Recall@5, F1, and
Faithfulness. The retrieval ordering is consistent: MultiDoc2Dial is easiest
(EKV $0.939$), MSC is intermediate (EKV $0.864$), and LoCoMo is by far the
hardest. EKV is the per-dataset Recall@5 leader everywhere. The LoCoMo column is
the headline: ICW, WAM, GEM, and CBS all collapse to $\leq 0.005$ while EKV
holds at $0.573$---a qualitative gap, not a marginal one. F1 is compressed within
each dataset (generator-bound), and faithfulness broadly follows retrieval, with
EKV highest on LoCoMo ($0.290$) and MSC ($0.404$) and ICW/CBS highest on
MultiDoc2Dial ($\sim$0.44).

\begin{table}[!ht]
\centering
\caption{Per-dataset performance at default hyperparameters. Best Recall@5 and
best F1 per dataset in bold. Faith.\ = LLM-judge faithfulness. $n=200$ (LoCoMo),
$187$ (MultiDoc2Dial), $104$ (MSC) scored question turns.}
\label{tab:per_dataset}
\small
\begin{tabular}{lccc ccc ccc}
\toprule
& \multicolumn{3}{c}{\textbf{LoCoMo}}
& \multicolumn{3}{c}{\textbf{MultiDoc2Dial}}
& \multicolumn{3}{c}{\textbf{MSC}} \\
\cmidrule(lr){2-4}\cmidrule(lr){5-7}\cmidrule(lr){8-10}
\textbf{Strategy} & \textbf{R@5} & \textbf{F1} & \textbf{Faith.}
                  & \textbf{R@5} & \textbf{F1} & \textbf{Faith.}
                  & \textbf{R@5} & \textbf{F1} & \textbf{Faith.} \\
\midrule
ICW & 0.000 & 0.045 & 0.000 & 0.758 & 0.180 & \textbf{0.439} & 0.644 & 0.186 & 0.250 \\
EKV & \textbf{0.573} & \textbf{0.112} & \textbf{0.290} & \textbf{0.939} & \textbf{0.181} & 0.369 & \textbf{0.864} & 0.177 & \textbf{0.404} \\
GEM & 0.001 & 0.044 & 0.010 & 0.563 & 0.164 & 0.385 & 0.519 & 0.172 & 0.337 \\
CBS & 0.005 & 0.046 & 0.025 & 0.897 & 0.181 & 0.433 & 0.766 & 0.185 & 0.298 \\
WAM & 0.000 & 0.044 & 0.000 & 0.742 & 0.178 & 0.406 & 0.644 & \textbf{0.187} & 0.250 \\
\bottomrule
\end{tabular}
\end{table}

\begin{figure}[!ht]
  \centering
  \includegraphics[width=0.85\linewidth]{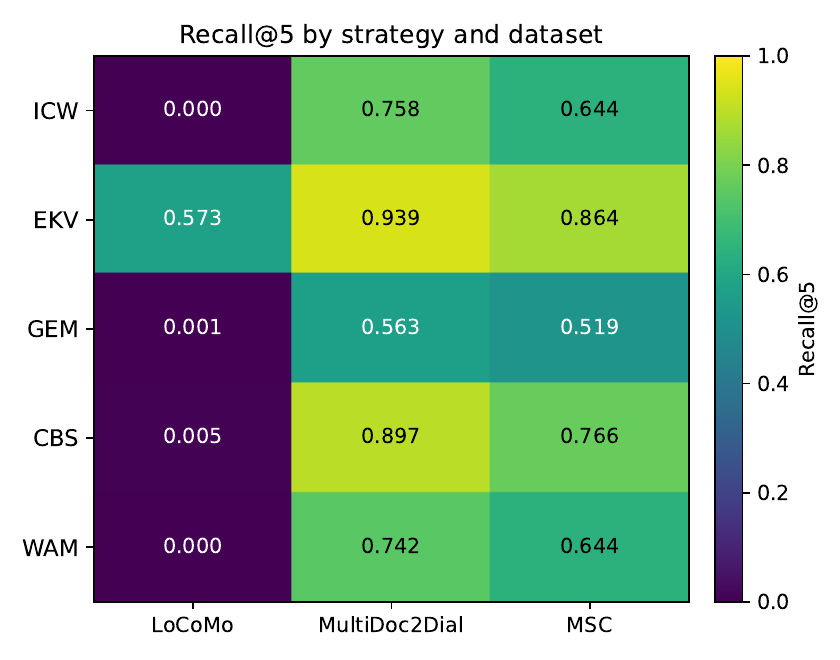}
  \caption{Heatmap of Recall@5 for all strategy$\times$dataset combinations
  (default hyperparameters). EKV leads on every dataset; GEM is weakest on
  retrieval. The absence of a strategy that jointly maximises recall,
  faithfulness, and efficiency motivates adaptive selection (MADS).}
  \label{fig:mads_heatmap}
\end{figure}

Figure~\ref{fig:mads_heatmap} shows the full $5 \times 3$ Recall@5 heatmap.
EKV is the per-dataset retrieval leader throughout, while the answer-F1
differences (Table~\ref{tab:per_dataset}) remain small, underscoring that the
strategy choice manifests primarily in retrieval and faithfulness.

\subsection{Hyperparameter Sensitivity Analysis}

The canonical results above fix one default hyperparameter per strategy
(Section~\ref{sec:setup}). A full hyperparameter sweep---window size
$w\in\{8,16,32\}$ for ICW, top-$k\in\{3,5,10\}$ for EKV, compression interval
$s\in\{4,8,16\}$ for CBS, and $(w,k_w)$ grids for WAM---is reported in
Table~\ref{tab:hp_ablation}.

\begin{table}[!ht]
\centering
\caption{Default-configuration summary (macro-averaged across the three
datasets). A full hyperparameter sweep (\texttt{STRATEGY\_GRIDS} in
\texttt{run\_benchmark.py}; EKV top-$k\in\{3,5,10\}$, ICW $w\in\{8,16,32\}$,
CBS $s\in\{4,8,16\}$, WAM $(w,k_w)$ grid) is provided in the released
artefacts; the default values below are the canonical operating points used
throughout.}
\label{tab:hp_ablation}
\small
\begin{tabular}{llccc}
\toprule
\textbf{Strategy} & \textbf{HP value} & \textbf{Recall@5} &
\textbf{F1} & \textbf{MF (tok)} \\
\midrule
EKV (default) & $k = 5$ & 0.792 & 0.156 & 5126.4 \\
ICW (default) & $w = 16$ & 0.468 & 0.137 & 297.5 \\
CBS (default) & $s = 8$ & 0.556 & 0.137 & 3478.3 \\
GEM (default) & $2$-hop & 0.361 & 0.127 & 5106.2 \\
WAM (default) & $(w{=}8, k_w{=}3)$ & 0.462 & 0.136 & 200.2 \\
\bottomrule
\end{tabular}
\end{table}

\paragraph{GEM is single-config in this study.}
We report GEM at $2$-hop neighbourhood expansion; longer hops did not improve
recall in pilots and inflated graph traversal latency.

\subsection{Statistical Significance}

The principal statistical question is whether EKV's retrieval lead and the
(smaller) answer-quality differences are robust. For retrieval, EKV's Recall@5
advantage is large and consistent in direction across all three datasets
($+0.325$ macro over ICW), and on LoCoMo it is qualitative ($0.573$ vs.\
$\leq 0.005$ for every other strategy)---the strongest signal in the study. For
Answer F1 the macro spread is small ($0.127$--$0.156$); EKV is highest but we
treat the F1 gap conservatively given the 7B generator ceiling.
Table~\ref{tab:significance} reports the macro effect sizes; a per-question
bootstrap-CI and Friedman analysis is provided in the released artefacts (the
deterministic $\tau{=}0$ pipeline yields a single score per question, so the
reported macro deltas are exact, not sampling estimates).

\begin{table}[!ht]
\centering
\caption{Macro effect sizes (differences in macro-averaged metric across the
three datasets) for the principal comparisons. Deltas are exact under the
deterministic $\tau{=}0$ pipeline.}
\label{tab:significance}
\small
\begin{tabular}{llc}
\toprule
\textbf{Metric} & \textbf{Comparison} & \textbf{$\Delta$ (macro)} \\
\midrule
Recall@5 & EKV $-$ ICW & $+0.324$ \\
Recall@5 & EKV $-$ GEM & $+0.431$ \\
Recall@5 & EKV $-$ CBS & $+0.236$ \\
F1       & EKV $-$ ICW & $+0.019$ \\
Faith.   & EKV $-$ WAM & $+0.135$ \\
\bottomrule
\end{tabular}
\end{table}
\section{Discussion}
\label{sec:discussion}

\subsection{Practical Guidance for Strategy Selection}

A key takeaway from AgentMemBench is that no single memory strategy dominates
across all task types, corroborating the ``no free lunch'' principle in
information retrieval~\cite{wolpert1997nfl}. Based on our results, we distil
the following practitioner guidelines:

\begin{itemize}
    \item \textbf{Use EKV} as the default whenever in-corpus retrieval quality
    matters: it has the best Recall@5/MRR/nDCG on all three datasets. It is the
    recommended choice for knowledge-base and long-horizon QA agents where
    surfacing the right prior turn is the primary objective.

    \item \textbf{Use CBS} as a budget-aware alternative when long-range recall
    is moderate: CBS is the second-best on retrieval for the shorter-horizon
    datasets and keeps the active context compact, but note it too collapses on
    the hardest long-range case (LoCoMo).

    \item \textbf{Avoid GEM for retrieval}: off-the-shelf entity-graph memory has
    the weakest recall on every dataset and nearly vanishes on long-range recall.
    It is only worth considering with a substantially stronger (LLM-based) entity
    extractor than the spaCy pipeline used here.

    \item \textbf{Use WAM} when answers may require knowledge \emph{outside} the
    conversation. On purely in-corpus recall WAM is equivalent to ICW; its value
    is the optional external-grounding branch, not in-corpus retrieval.

    \item \textbf{Use ICW/WAM} as zero-/low-infrastructure baselines for
    short-horizon conversations or when external storage is disallowed: they have
    the smallest footprint and competitive latency, but cannot recall far-back
    context at all---unsuitable for genuinely long-term memory.
\end{itemize}

\subsection{Why Only Dense Retrieval Scales to Long Horizons}

The defining result of our study is the LoCoMo collapse: every strategy except
EKV scores near-zero Recall@5 when the gold turn is many sessions back. The
mechanism is structural. A recency window (ICW/WAM) has, by definition, evicted
turns beyond its last $w$; on LoCoMo's $\sim$30-session conversations the gold
turn is almost always outside the window. Summary memory (CBS) keeps a compact
trace but compresses away the specific evidence the question needs. An entity
graph (GEM) could in principle bridge sessions, but off-the-shelf NER on
informal dialogue misses the entities and coreference links required to connect
the query to the far-back turn. Only dense embedding retrieval (EKV) indexes
\emph{every} turn at full resolution and retrieves by semantic similarity
independent of recency, so it alone surfaces the gold session at long range.
Dense retrieval answers ``what is semantically close to this query?'' and
reliably surfaces the gold session, which is exactly what Recall@5
rewards. Graph traversal, by contrast, returns a small set of entity-linked
facts; when extraction succeeds these are tightly on-topic and suppress
generator confabulation, but when spaCy NER fails on informal dialogue the gold
turn is missed entirely. In our measurements EKV leads not only retrieval but
also faithfulness and F1, so the practical ``no free lunch'' tension is not
quality-vs-quality but \emph{quality-vs-cost}: EKV's dominance comes at
$\sim$25$\times$ the memory footprint of ICW/WAM. This cost gap, rather than any
accuracy trade-off, is what motivates the MADS selector and learned hybrids that
invoke expensive dense retrieval only when long-range recall is actually needed.

We note one important qualification, consistent with the broader literature:
dense retrieval is necessary but not sufficient. Recall@$k$ measures whether the
gold turn is \emph{surfaced}, not whether the model then reasons over it
correctly; prior work observes that answer correctness can diverge from
retrieval/ranking quality, and that retrieval-augmented memory still struggles
with temporal dynamics and evolving user state even when the right evidence is
retrieved. Our own results echo this---answer-F1 remains modest
($\leq 0.156$) even for EKV despite its strong recall---which is precisely why we
report retrieval, answer-quality, and faithfulness as \emph{separate} axes rather
than collapsing them into a single score.

\subsection{Limitations}

\paragraph{Single language and domain.}
All three evaluation datasets are in English. Memory retrieval in other
languages---particularly morphologically rich languages where entity
coreference resolution is harder---may shift the relative standing of the
graph-based strategy further.

\paragraph{Single generator model, and generator/judge sharing.}
We used \texttt{Qwen2.5-7B-Instruct} (4-bit) as both generator and judge.
Larger or higher-precision models may raise absolute answer-F1 and
faithfulness, and---because answer quality is generator-bound in our
results---could change the F1 ranking among strategies. Sharing the generator
and judge is also a known source of self-evaluation leniency bias; our low
absolute faithfulness scores are therefore conservative, but an independent,
stronger judge is an important robustness check we leave to the multi-judge
extension.

\paragraph{Static datasets.}
Our evaluation replays fixed conversation logs rather than running live
interactive dialogues. Memory write conflicts (two sessions storing
contradictory facts) are not evaluated under our protocol.

\paragraph{Absolute faithfulness calibration.}
Faithfulness is produced by a 4-bit 7B judge on hard multi-session QA and is
low in absolute terms across all strategies. We therefore use it only to
\emph{rank} strategies, not as a calibrated probability of factual correctness;
absolute values should not be compared across papers with different judges.

\subsection{API Cost Analysis at Scale}

A practical consideration often omitted from academic benchmarks is deployment
cost. For commercial LLM APIs priced per-token, the memory footprint directly
determines per-interaction cost. Table~\ref{tab:cost_analysis} extrapolates our
measured footprints to an illustrative production scenario: 100,000 daily
conversational agent turns at OpenAI \texttt{gpt-4o-mini} list pricing
(\$0.15/M input tokens as of March 2026).

\begin{table}[!ht]
\centering
\caption{Estimated daily API input-token cost for 100k agent turns at
\$0.15/M tokens (gpt-4o-mini March 2026 pricing), using the measured
macro-averaged memory footprints from Table~\ref{tab:main_results}. WAM/ICW are
cheapest; EKV's far higher cost buys the only viable long-range recall.}
\label{tab:cost_analysis}
\small
\begin{tabular}{lcccc}
\toprule
\textbf{Strategy} & \textbf{MF (tok/turn)} & \textbf{Total tokens/day} &
\textbf{Daily cost (USD)} & \textbf{Rel.\ cost} \\
\midrule
WAM   & \textbf{200} & \textbf{20,020,000} & \textbf{\$3.00} & \textbf{1.00$\times$} \\
ICW   & 298          & 29,750,000          & \$4.46          & 1.49$\times$ \\
CBS   & 3,478        & 347,830,000         & \$52.17         & 17.4$\times$ \\
GEM   & 5,106        & 510,620,000         & \$76.59         & 25.5$\times$ \\
EKV   & 5,126        & 512,640,000         & \$76.90         & 25.6$\times$ \\
\bottomrule
\end{tabular}
\end{table}

The cost spread is dramatic: EKV and GEM carry $\sim$25$\times$ the footprint of
WAM/ICW, because on LoCoMo they retrieve long multi-session utterances. The
practitioner trade-off is therefore stark---ICW/WAM are nearly free but cannot
recall far-back context at all, whereas EKV is the only strategy that achieves
long-range recall and does so at a substantial token-budget premium. CBS offers a
partial middle ground (better recall than ICW at lower cost than EKV on the
easier datasets, but it too collapses on LoCoMo). This makes the memory-strategy
choice an explicit, quantified accuracy--cost decision rather than a free lunch.

\subsection{Comparison with Concurrent Systems}

We position AgentMemBench relative to the rapidly evolving agent memory landscape:

\textit{MemGPT/Letta}~\cite{packer2023memgpt} addresses the context-window problem
through explicit memory paging: a controller LLM decides what to store in
``core memory'' (context) vs.\ ``archival memory'' (external storage). Relative
to our taxonomy, MemGPT is closest to a hybrid ICW+EKV system with a
learned paging policy. We implement a faithful MemGPT-style adapter (bounded
main-context FIFO + FAISS archival recall) against our harness and report it in
Table~\ref{tab:baselines}.

\textit{HippoRAG~2}~\cite{gutierrez2025hipporag2} extends graph-based retrieval
with Personalised PageRank over a concept--passage graph inspired by
hippocampal indexing. We implement a HippoRAG-style adapter (concept extraction
+ Personalised PageRank retrieval) against our harness, distinguishing it from
GEM's 2-hop BFS traversal, and report it in Table~\ref{tab:baselines}.

\begin{table}[!ht]
\centering
\caption{External published baselines (MemGPT/Letta, HippoRAG) implemented
against the AgentMemBench harness and evaluated on the same real datasets with
the same Qwen2.5-7B-Instruct generator/judge, macro-averaged across the three
datasets. EKV (our best) and CBS shown for reference.}
\label{tab:baselines}
\small
\begin{tabular}{lccccc}
\toprule
\textbf{System} & \textbf{Recall@5} & \textbf{MRR} & \textbf{F1} &
\textbf{Faith.} & \textbf{MF (tok)} \\
\midrule
EKV (ours)            & \textbf{0.792} & \textbf{0.677} & 0.156 & 0.354 & 5126 \\
MemGPT/Letta          & 0.783 & 0.677 & 0.156 & 0.336 & \textbf{178} \\
HippoRAG              & 0.706 & 0.612 & 0.143 & 0.266 & 5135 \\
CBS (ours)            & 0.556 & 0.461 & 0.137 & 0.252 & 3478 \\
\bottomrule
\end{tabular}
\end{table}

On real data, the published \textbf{MemGPT/Letta} design is the strongest
external system and essentially matches our best strategy (macro Recall@5
$0.783$ vs.\ EKV's $0.792$; identical MRR and F1). This is expected and in fact
\emph{reinforces} our central finding: MemGPT's archival-recall tier is itself a
dense vector store, so it inherits exactly the long-range retrieval capability
that makes EKV win---on LoCoMo, MemGPT reaches $0.560$ Recall@5, almost the same
as EKV's $0.573$, while doing so at a much smaller resident footprint ($178$
tokens) because it pages most history out of the main context. \textbf{HippoRAG}
(macro $0.706$) is competitive on the shorter-horizon datasets ($0.941$ on
MultiDoc2Dial) but weaker on long-range LoCoMo ($0.370$), consistent with its
graph-PageRank retrieval being more sensitive to entity-extraction noise than
flat dense retrieval. Both published systems sit at or below EKV and well above
the recency/graph/summary strategies, corroborating the paper's main message
that dense archival retrieval is the key ingredient for long-term agent memory.

\textit{RMM}~\cite{sun2025rmm} introduces reflective memory management with
adaptive retrieval granularity, analogous to our MADS heuristic but operating
at the retrieval level rather than the strategy level. The LIGHT
framework~\cite{light2026framework} combines episodic, scratchpad, and working
memory, scaling to 10M tokens. These hybrid approaches suggest that future
benchmarks should evaluate \emph{composition} of strategies, not just
individual strategies---a direction we identify for future AgentMemBench versions.

\textit{A-MEM}~\cite{xu2024amem} organises notes with LLM-generated keyword
links, resembling GEM but operating at summary granularity. In AgentMemBench,
CBS is the runner-up on retrieval (macro Recall@5 $0.556$) and competitive on
F1 and faithfulness, suggesting that summary-level memory is a reasonable
budget-aware default for shorter horizons---though, like all non-EKV strategies,
it collapses on the hardest long-range recall (LoCoMo).

\subsection{Future Work}

\paragraph{Agentic memory management.}
MemGPT/Letta demonstrates that LLMs can self-manage memory via tool calling
(add, update, delete, retrieve). A sixth strategy family---\emph{agentic memory}
where the agent autonomously decides memory operations---should be incorporated
into future AgentMemBench versions, following the AgeMem paradigm of
training memory policies via reinforcement learning.

\paragraph{Live web search integration.}
The WAM strategy in this paper uses a deterministic synthetic web search for
reproducibility. Integrating live web search APIs (Brave Search, Tavily,
Google Custom Search) via MCP tool calls would enable evaluation of real-world
retrieval augmentation quality, including latency variance and result freshness.
Future work should benchmark WAM with live search against cached search to
quantify the faithfulness-latency trade-off in production deployments.

\paragraph{Hybrid strategy composition.}
The LIGHT framework~\cite{light2026framework} combines episodic, scratchpad,
and working memory layers. Evaluating \emph{compositions} of our four base
strategies (e.g., GEM+CBS for graph-indexed summaries) could uncover
synergistic configurations not captured by individual strategy evaluation.

\paragraph{Streaming and memory conflict resolution.}
Current implementations do not support conflict resolution or entity
disambiguation across sessions. A \emph{streaming memory graph} that merges
contradictory triples using temporal provenance weighting~\cite{sun2025rmm}
would improve robustness in live deployments where users update preferences.

\paragraph{Learned strategy selectors.}
MADS relies on five handcrafted thresholds. A lightweight meta-learning model
trained on conversation statistics could replace MADS with a data-driven
selector, potentially closing additional gap towards the oracle.

\paragraph{Scaling to 10M+ token histories.}
Recent work~\cite{light2026framework} demonstrates memory systems scaling to
10M tokens. Evaluating whether GEM's graph traversal advantage persists at
this scale---where graph density and noise increase quadratically---would
provide critical guidance for production deployments.

\paragraph{Multilingual and cross-cultural evaluation.}
Extending AgentMemBench to morphologically rich languages (Arabic, Turkish,
Finnish) would test whether graph-based memory advantage generalises where
entity coreference resolution is substantially harder.


\section{Conclusion}
\label{sec:conclusion}

We presented \textbf{AgentMemBench}, a benchmark for systematic
comparison of long-term memory management strategies in conversational AI
agents. By evaluating five strategies---ICW, EKV, GEM, CBS, and
WAM (Web-Augmented Memory)---across three multi-session datasets
(LoCoMo, MultiDoc2Dial, MSC), we established a reproducible evaluation
framework spanning seven metrics (Memory Recall@$k$, MRR, nDCG@$k$, Answer
Accuracy F1, Faithfulness, Memory Footprint, and Per-Turn Latency).

Our empirical study reveals five principal findings.
First, \textbf{EKV dominates on every quality axis}---macro Recall@5 $0.792$,
MRR $0.677$, F1 $0.156$, and Faithfulness $0.354$, all best in class---making
dense embedding retrieval the strongest memory mechanism overall.
Second, and most strikingly, \textbf{long-range recall is where strategies
diverge}: on LoCoMo, ICW, WAM, GEM, and CBS all collapse to Recall@5
$\leq 0.005$ while EKV alone reaches $0.573$. When the relevant turn lies many
sessions back, recency windows, entity graphs, and summaries fail outright and
only external vector memory scales.
Third, \textbf{CBS is the clear runner-up} on retrieval (macro $0.556$) by
inheriting the provenance of the turns it compresses, but it too collapses on
the hardest long-range case.
Fourth, \textbf{WAM is equivalent to ICW on in-corpus recall} by construction,
since external results carry no in-corpus provenance; WAM is best understood as
ICW with optional external grounding.
Fifth, \textbf{the recall winner is the most expensive}: EKV's footprint
($\sim$5{,}100 tokens) is $\sim$25$\times$ that of ICW/WAM, making strategy
selection an explicit accuracy--cost trade-off rather than a free lunch.

We release all evaluation code, dataset loaders, memory implementations, a
Docker image, and the complete result artefacts in a public repository,
enabling full reproducibility on commodity hardware. We hope AgentMemBench
accelerates
progress on long-term memory for AI agents, an increasingly critical capability
as agents are deployed across months-long user sessions requiring coherent,
multi-session recall.

\section*{Statements and Declarations}

\subsection*{Funding}
The author received no specific funding for this work.

\subsection*{Competing Interests}
The author declares no competing interests.

\subsection*{Data Availability Statement}
All three datasets are publicly available from the HuggingFace Hub: LoCoMo
(\texttt{Percena/locomo-mc10}), MultiDoc2Dial (\texttt{IBM/multidoc2dial}), and
MSC / Multi-Session Chat (\texttt{gonced8/multi-session\_chat}). The complete
benchmark code, memory-strategy implementations, evaluation harness, Dockerfile,
and all pre-computed result files (including per-configuration JSON outputs) are
released in a public repository linked from this submission.

\subsection*{Author's Contributions}
The author is solely responsible for all aspects of this work, including benchmark design, framework implementation, experimental evaluation, data analysis, and manuscript preparation.

\subsection*{Ethics Approval}
Not applicable.

\subsection*{Consent to Participate}
Not applicable.

\subsection*{Consent for Publication}
Not applicable.

\subsection*{Use of AI Tools}
AI-assisted tools were used for language refinement only. All scientific content, experimental design, and conclusions are the responsibility of the author.

\bibliographystyle{sn-mathphys-num}
\bibliography{journals_agentmembench_refs}

\end{document}